\documentclass{ieeeaccess}
\usepackage{kotex}
\usepackage{cite}
\usepackage{booktabs}
\usepackage{multirow}
\usepackage{multicol}
\usepackage{bm}
\usepackage{graphicx}
\usepackage{colortbl}
\usepackage{url}
\usepackage[normalem]{ulem}
\usepackage[table,xcdraw]{xcolor}
\useunder{\uline}{\ul}{}
\usepackage{amsmath}
\usepackage{amssymb,amsfonts}
\usepackage{algorithmic}
\usepackage{textcomp}
\def\BibTeX{{\rm B\kern-.05em{\sc i\kern-.025em b}\kern-.08em
    T\kern-.1667em\lower.7ex\hbox{E}\kern-.125emX}}
\begin{document}
\history{Date of publication xxxx 00, 0000, date of current version xxxx 00, 0000.}
\doi{10.1109/ACCESS.2023.0322000}

\title{DSG-KD: Knowledge Distillation from Domain-Specific to General Language Models}
\author{
\uppercase{Sangyeon Cho}\authorrefmark{1}, \IEEEmembership{Graduate Student Member, IEEE},
\uppercase{Jangyeong Jeon}\authorrefmark{1},
\uppercase{Dongjoon Lee}\authorrefmark{1},
\uppercase{Changhee Lee}\authorrefmark{1}
and \uppercase{Junyeong Kim}\authorrefmark{1} \IEEEmembership{Member, IEEE}
}
\address[1]{Department of Artificial Intelligence, Chung-Ang University, Seoul 06974, Republic of Korea}
\tfootnote{This work was partly supported by Institute of Information \& communications Technology Planning \& Evaluation (IITP) grant funded by the Korea government(MSIT) (No.2022-0-00184, Development and Study of AI Technologies to Inexpensively Conform to Evolving Policy on Ethics) and partly supported by Institute of Information \& communications Technology Planning \& Evaluation (IITP) grant funded by the Korea government(MSIT) (No.2021-0-01341, Artificial Intelligence Graduate School Program, Chung-Ang University)}

\markboth
{S. Cho \headeretal: DSG-KD: Knowledge Distillation from Domain-Specific to General Language Models}
{S. Cho \headeretal: DSG-KD: Knowledge Distillation from Domain-Specific to General Language Models}

\corresp{Corresponding author: Junyeong Kim (e-mail: junyeongkim@cau.ac.kr).}

\begin{abstract}
The use of pre-trained language models fine-tuned to address specific downstream tasks is a common approach in natural language processing (NLP). However, acquiring domain-specific knowledge via fine-tuning is challenging. Traditional methods involve pretraining language models using vast amounts of domain-specific data before fine-tuning for particular tasks. This study investigates emergency/non-emergency classification tasks based on electronic medical record (EMR) data obtained from pediatric emergency departments (PEDs) in Korea. Our findings reveal that existing domain-specific pre-trained language models underperform compared to general language models in handling N-lingual free-text data characteristics of non-English-speaking regions. To address these limitations, we propose a domain knowledge transfer methodology that leverages knowledge distillation to infuse general language models with domain-specific knowledge via fine-tuning. This study demonstrates the effective transfer of specialized knowledge between models by defining a general language model as the student model and a domain-specific pre-trained model as the teacher model. In particular, we address the complexities of EMR data obtained from PEDs in non-English-speaking regions, such as Korea, and demonstrate that the proposed method enhances classification performance in such contexts. The proposed methodology not only outperforms baseline models on Korean PED EMR data, but also promises broader applicability in various professional and technical domains. In future works, we intend to extend this methodology to include diverse non-English-speaking regions and address additional downstream tasks, with the aim of developing advanced model architectures using state-of-the-art KD techniques.
The code is available in \url{https://github.com/JoSangYeon/DSG-KD}.
\end{abstract}


\titlepgskip=-21pt

\maketitle
\section{Introduction}
\label{sec:introduction}

\begin{figure}[t]
\centering
\includegraphics[width=\columnwidth]{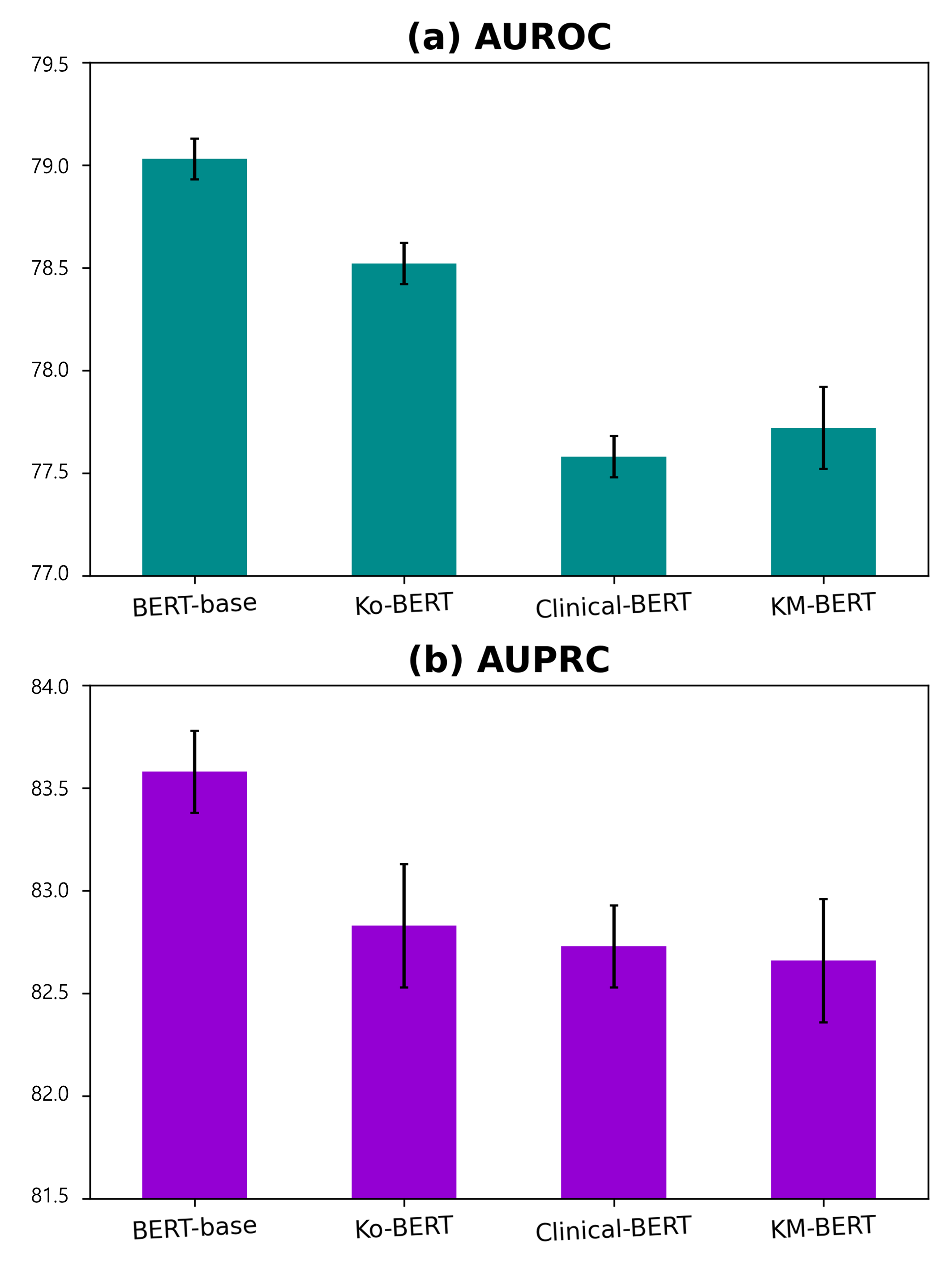} 
\caption[Generalized language model (LM) performs better.]{Generalized LM performs better. Performance of each pre-trained LM on an emergency/non-emergency classification task using EMR data from Korean PEDs in terms of (a) AUROC and (b) AUPRC.}
\label{fig:lm_perform}
\end{figure}

\PARstart{W}{i}th the proactive utilization of Electronic Medical Records (EMR) by healthcare institutions worldwide, a vast amount of medical information is being stored as data. \cite{menachemi2011benefits,hartswood2003making,williams2008role} In particular, EMRs are documented in Pediatric Emergency Departments (PEDs) to capture patients' conditions at the time of visit, test results, and additional details in the form of free-text. These free-text entries are crucial components of EMRs and include essential clinical notes, such as the patient's gender, age, and vital signs. Given the importance of initial responses in PEDs, distinguishing between emergency and non-emergency patients based on EMRs recorded at the time of admission is critical. \cite{hong2018predicting,lundberg2017unified,lambert2017paediatric} However, owing to the nature of PED environments, clinical notes are often recorded in an unstructured and inconsistent manner, complicating the classification of emergency and non-emergency patients based on them. \cite{harrison2021machine,zhang2019high,castro2015validation} This reduces the utility of the data in supporting decision-making by healthcare professionals, such as physicians. \cite{adnan2020role, tayefi2021challenges}

In particular, EMR data obtained from PEDs in Korea, a non-English-speaking country, are written in Korean. Non-English-speaking countries present an additional layer of complexity in classification tasks owing to the use of both English and local languages. \cite{gianfrancesco2021narrative,barbazza2021current} In such cases, medical jargon and test results are often written in English or using English abbreviations, while the patient's condition and chief complaints are predominantly documented in the local language (Korean). \cite{park2017current,shinozaki2020electronic} Thus, in non-English speaking countries, the data often come in the form of N-lingual free-text, with critical medical terms presented in English and other parts present in the local language. This form of data presents significant challenges in downstream tasks and during the extraction of meaningful words.

To overcome these problems, language models have been introduced based on transformers pre-trained on specific domain data to equip them with general domain knowledge. \cite{rasmy2021med, alsentzer2019publicly} In particular, KM-BERT \cite{kim2022pre}, pre-trained on Korean medical data, has been proposed to perform downstream tasks in the medical domain based on fine-tuning. However, KM-BERT does not perform well in the context of data collected in non-English-speaking countries, such as Korea. As depicted in Figure~\ref{fig:lm_perform}, in the task of classifying emergency and non-emergency cases using EMR data obtained from Korean PEDs, general language models, such as Ko-BERT and BERT-base, outperformed domain-specific pre-trained models, such as KM-BERT and Clinical-BERT, in terms of Area Under the Receiving Operating Curve (AUROC) and Area Under the Precision-Recall Curve (AUPRC) \cite{lasko2005use,flach2015precision}. This indicates that language models pre-trained with medical domain data are vulnerable to complications caused by N-lingual free-text characteristics of EMR data. The current study was envisioned to address this issue.

In this study, we address this problem by extracting domain knowledge for training using Knowledge Distillation (KD) \cite{hinton2015distilling}. We define an LM pre-trained with medical domain data as the teacher model and a general LM as the student model. Then, we extract medical knowledge from the former model and transfer it to the general latter model. We identify words containing medical knowledge in the input text and design the training to enable the student model to learn the teacher model’s hidden states and attention matrices for these words. Our experiments demonstrate that the proposed method achieves effective knowledge transfer between LMs, yielding the best performance in classifying emergency and non-emergency cases based on PED EMR data.

\begin{table*}[!ht]
\centering
\resizebox{0.7\textwidth}{!}{%
\begin{tabular}{@{}ccccc@{}}
\toprule
\multicolumn{1}{l}{} &
  \begin{tabular}[c]{@{}c@{}}\textbf{Total number} \\ \textbf{of words}\end{tabular} &
  \begin{tabular}[c]{@{}c@{}}\textbf{Number of Korean} \\ \textbf{words(medical words)}\end{tabular} &
  \begin{tabular}[c]{@{}c@{}}\textbf{Number of English} \\ \textbf{words(medical words)}\end{tabular} &
  \begin{tabular}[c]{@{}c@{}}\textbf{Number of} \\ \textbf{other words}\end{tabular} \\ \midrule
\textbf{Train} & 4,599,471 & 1,989,252 (99,197)   & 1,071,523 (220,221) & 1,538,696 \\
\textbf{Dev}   & 1,152,777 & 499,117 (24,590)     & 269,161 (55,156)    & 384,499   \\
\textbf{Test}  & 1,444,904 & 624,343 (31,372)     & 335,464 (68,684)    & 485,097   \\
\textbf{total} & \textbf{7,197,152} & \textbf{3,112,712 (155,159)} & \textbf{1,676,148 (344,061)} & \textbf{2,408,292} \\ \bottomrule
\end{tabular}%
}
\caption{Data Analysis for words: For the analysis of data on a word-by-word basis, the ratio of Korean:English:Other is used to signify the language distribution. In this case, it is approximately 0.43:0.23:0.33. The proportion of medical terms in Korean is 5\%, while that in English is 20\%, which is very high.}
\label{tab:txt_analysis}
\end{table*}

\section{Related Works}
\label{sec:Related_work}

\subsection{NLP for Clinical notes (EMR)}
EMR \cite{sheikhalishahi2019natural,ben2014ehr} represents digitally formatted data that are systematically collected and electronically stored to document patients' health information. In recent years, EMR data obtained from PEDs have garnered significant attention as the effective processing of such data is likely to enhance decision-making in clinical environments. \cite{hong2018predicting, anderson2015using, kirubarajan2020artificial} Various related studies have focused on the utilization, analysis, and decision-support methods of EMR data obtained from PEDs from multiple perspectives. \cite{weng2017medical,spasicclinical,khanday2020machine}

Early research on EMR data processing primarily targeted English-based data \cite{aman2007identifying}, applying traditional NLP methods such as Bag of Words (BoW) \cite{zuccon2013automatic,harpaz2014text,delahanty2019development} and Term Frequency-Inverse Document Frequency (TF-IDF) \cite{sparck1972statistical,salton1988term}. However, owing to the significant increase in the amount of acquired medical data in recent years, pre-trained LMs based on transformers have emerged \cite{medrouk2017deep, wu2020deep}, and active research has been conducted on fine-tuning these models for various medical datasets. Notably, LMs such as KM-BERT and Clinical-BERT have been pre-trained on extensive medical data, exhibiting great potential in various NLP tasks in the medical domain. \cite{rasmy2021med, yu2019biobert}

However, the aforementioned LMs exhibit certain vulnerabilities in the case of N-lingual, free-text-formatted EMR data obtained from PED environments of non-English-speaking countries (see Figure~\ref{fig:lm_perform}). Given the importance of initial responses in PEDs, such issues are critical. \cite{fuchs2016definitions, tohira2022machine} This study describes methodologies capable of effectively learning and processing such N-lingual and free-text formatted data.

\subsection{N-Lingual free-text data}
The processing of N-lingual and free-text data presents various motivations and challenges for NLP and deep learning. Early research on NLP focused primarily on English-based datasets. \cite{aman2007identifying} However, data obtained from diverse countries has motivated the study of N-lingual processing \cite{park1990korean,miwa1985intrasentential,ahn2017language,amazouz2017addressing}, leading to extensive research. This has led to the development of robust models capable of understanding the nuances of mixed languages to enhance the efficiency and accuracy of analyzing texts containing multiple languages \cite{pires2019multilingual}. Moreover, with the increase in online content, including social media posts, emails, news articles, and comments, the amount of unstructured free-text data has also increased. \cite{ahn2017language, baker2011foundations} Consequently, numerous studies on NLP have aimed to extract meaningful insights from unstructured data and transform them into structured information. In particular, LM development in NLP currently focuses on maximizing the understanding of free text. \cite{lee2020kcbert}

In non-English-speaking countries, the collection of various data types has led to the emergence of datasets that exhibit both N-lingual and free-text characteristics. \cite{chen2020cross, jin2019xlore2} This also constitutes a major topic of study within NLP. Indeed, research has been conducted on EMR data collected at PEDs in non-English-speaking countries, in which both N-lingual and free-text characteristics are prevalent. \cite{bae2021keyword, park2021framework} Existing research has identified critical issues with the performance of traditional LMs and NLP methods on such data. Unlike previous studies, we aim to enhance our understanding of data with N-lingual and free-text characteristics using KD.

\subsection{knowledge distillation}
In this study, domain knowledge is transferred from the teacher model to the student model using KD. \cite{hinton2015distilling} Traditionally, KD is used in the NLP field to transfer knowledge from a larger teacher model to a smaller student model, with a focus on model size and lightweighting. \cite{sanh2019distilbert, jiao2019tinybert} Previous research approached this by having the student model mimic the teacher model's predictions using relatively simple methods. Recently, KD between transformer-based LMs has evolved to mimic the representation vectors of hidden states and attention matrices. This study, however, does not focus on model size and lightweighting, but rather on "extracting the unique knowledge possessed by the teacher model and transferring it to the student model." Although we use conventional distillation methods \cite{sanh2019distilbert, jiao2019tinybert}, we first define the knowledge that the student model needs to learn and subsequently extract it from the teacher model for distillation. Detailed explanations are provided in [Section 3].

By examining previous studies that applied KD from a knowledge transfer perspective rather than a lightweighting perspective, active research has been conducted on transferring knowledge from image models to language models \cite{qin2021efficient,zhao2023mskd}; however, little research has been conducted on knowledge transfer between language models. \cite{guzman2023hybrid} Existing studies on knowledge transfer between language models have focused on the quality of overall knowledge transfer, rather than domain-specific knowledge transfer. Our approach differs from those proposed in previous studies in that it focuses on the exchange of domain knowledge between language models in different domains.

\begin{figure*}[!t]
\centering
\includegraphics[width=\textwidth]{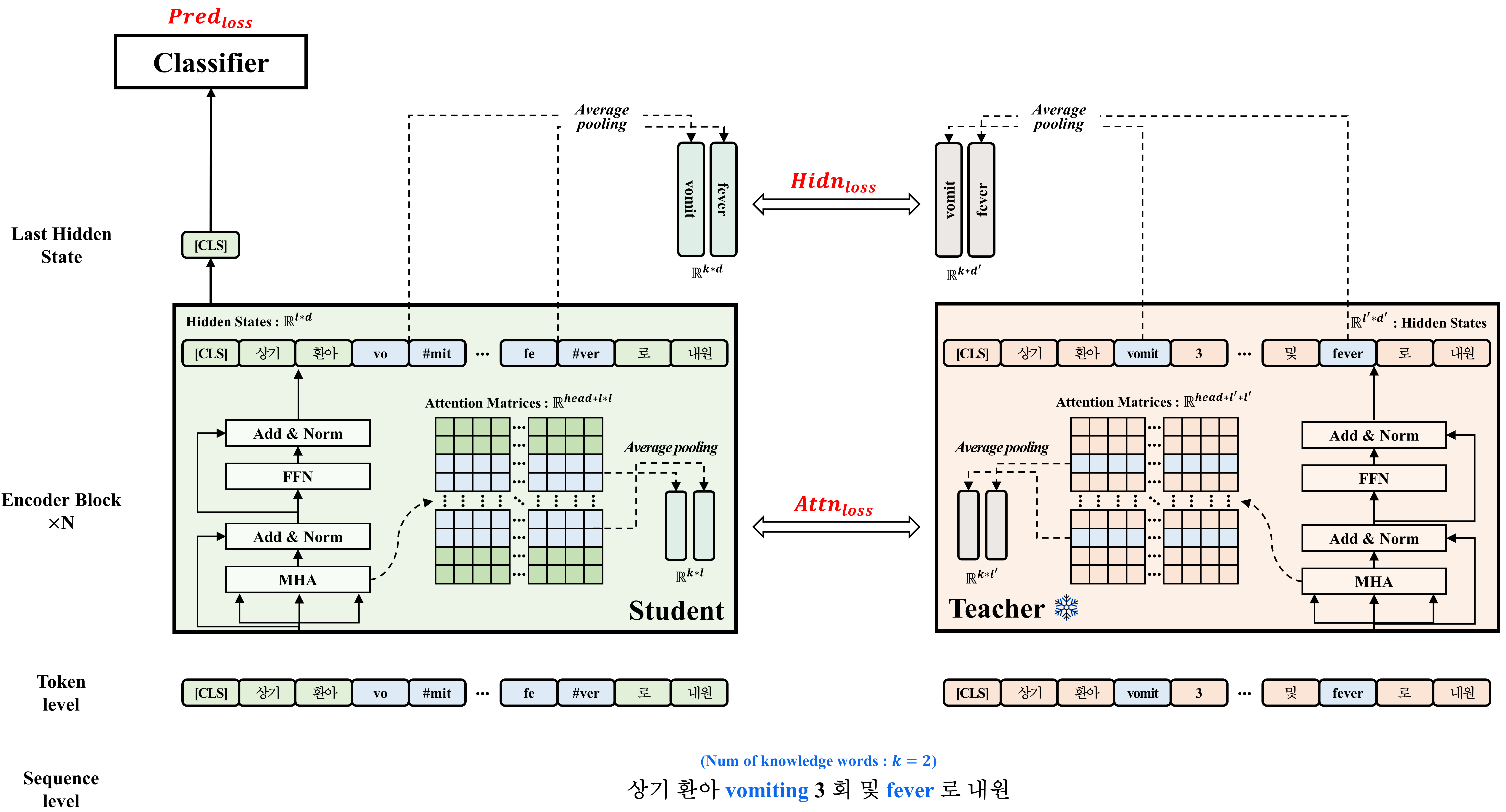} 
\caption{Model Architecture (Knowledge Transfer) - Visualization of the proposed methodology. The architecture consists of both student and teacher models based on transformers, which act as encoder blocks with multi-head attention (MHA) and feed-forward networks (FFN). The student model performs a prediction task ($\mathcal{L}_{\text{pred}}$) and receives appropriate domain knowledge from the teacher model by minimizing $\mathcal{L}_{\text{hidn}}$ and $\mathcal{L}_{\text{attn}}$. In the figure, we define "vomiting" and "fever" as domain knowledge words ($k=2$) and perform distillation by receiving appropriate representations from the hidden states and attention matrices that arise from the teacher model. Overall, the goal of the proposed architecture is to transfer the teacher model's knowledge to the student, not only in terms of classification predictions but also throughout the model's internal representation, thereby enabling the student model to make decisions based on a deeper and more nuanced understanding of the input data.}
\label{fig:architecture}
\end{figure*}

\section{Proposed Method}
\label{sec:method}
In this section, we formalize the central problem considered in this study and introduce the proposed methodology.

\subsection{Text Preprocessing \& Labeling}
EMR data are obtained from Korean PEDs and then analyzed and preprocessed before model training. Even though South Korea is a non-English-speaking country, English is often used alongside local languages. This is reflected in the data obtained, which are recorded in the form of bilingual free-text clinical notes. The preprocessing and analysis methods, described later, are summarized in Table \ref{tab:txt_analysis}.

First, the bilingual free-text clinical notes are preprocessed by performing language and symbol distinction processing and new character removal while retaining the meaning of the existing content as far as possible. Newline characters ($\backslash r$, $\backslash n$) generated during the process of recording and storing data in medical institutions are removed, and blanks are inserted between languages and symbols to separate them appropriately. This step is also performed for a method described below. Because we use a byte-pair encoding-based tokenizer, inserted blanks do not affect the input words severely\cite{berglund2023formalizing}. In addition, because it is not possible to specify the features to be used as labels in the acquired data, active doctors are consulted to determine the label criteria using columns such as test status and medication history, as recorded in the data.

As per the preprocessed data, approximately 43\% of the words are Korean, 23\% are English, and 33\% are symbols and other words. In addition, the number of English words in the data that are medical terms is evaluated by creating our own Korean medical dictionary by crawling the Korean Medical Search Engine and online medical dictionaries. According to Table \ref{tab:txt_analysis}, approximately 20\% of the English words in the data are medical words, whereas only 5\% of the Korean words are medical terms. This implies that almost all medical terms are more biased toward English than local languages and that medical domain knowledge is more embedded in English than in local languages.

\subsection{Method}
This section describes the domain knowledge transfer methodology proposed in this thesis in detail. The goal of this methodology is to transfer domain knowledge from a language model (teacher) pre-trained on a specific domain to a language model (student) pre-trained on general-purpose data. An overview of the framework is presented in Figure ~\ref{fig:architecture}.

We define $S$ as a generalized language model that categorizes urgent/non-urgent cases based on free-text clinical note data as input data. We train and validate the student model $S$. In addition, we define a domain-specific language model $T$ that conveys domain knowledge and aids learning when $S$ performs emergency or non-emergency classification tasks. The input free-text clinical notes are denoted by $X=\{x_i^S, x_i^T\}_{i=0}^N$. Here, $x_i^S$ denotes the input of the $S$ model, and $x_i^T$ denotes the input of the $T$ model. We also define $Y=\{y_i\}_{i=0}^N$, $Y \in \{0, 1\}$, and label it as emergency/non-emergency. Finally, we construct a dataset $\text{Data} = \{X, Y\}$ consisting of $X$ and $Y$.

\subsubsection{Definition of Domain Knowledge}
To transfer domain knowledge, we must define the domain knowledge present in the input $x_i$. This is defined as English words in the PED's EMR data (because, as discussed previously, medical knowledge that the teacher needs to extract is almost always expressed in English, instead of the local language). To represent domain knowledge words (subwords) in the input sequence $x_i$, we represent the knowledge mask $\mathbf{M}$ as follows:
\begin{equation}
    \label{eq. domain_knowledge_mask}
    \begin{aligned}
        x_i &= \{t_1, t_2, \cdots, t_{l-1}, t_l\} \in \mathbb{R}^{l} \\
        \mathbf{M}_i &= \{m_1, m_2, \cdots, m_{l-1}, m_l\} \in \{0, 1, \cdots, k\}^{l}
    \end{aligned}
\end{equation}
where the knowledge mask $\mathbf{M}$ is $\mathbf{M} \in \mathbb{R}^{l}$, where each element $m_l$ of $\mathbf{M}$ represents the index of the word for which the token at that position expresses a particular piece of domain knowledge. If $m_l = 0$, the token does not contain domain knowledge; if $m_l > 0$, the input $x_i$ contains domain knowledge. In addition, $k$ denotes the number of domain knowledge words. The appropriate domain knowledge token $d_i$ is extracted from the input sequence $x_i$ using $\mathbf{M}$.
\begin{equation}
    \label{eq. domain_knowledge_tokens}
    \begin{aligned}
        d_i = \{x_i[j] | m_j > 0, \forall j \in \{0, \cdots, l\}\}
    \end{aligned}
\end{equation}
where $d_i$ denotes the domain knowledge tokens corresponding to each token of the input $x_i$ with the set relation $d_i \in x_i$. 

Choosing the scope of domain knowledge definition is directly linked to model performance in the proposed methodology. In the data considered in this study, knowledge words are not defined as exclusively medical terms because the boundaries between medical and non-medical terms in the language are blurred. Thus, embedding the medical background knowledge of the teacher model  in the English language itself is expected to benefit student learning. In particular, defining the appropriate domain knowledge based on the available data can be adopted as an inductive bias in the proposed domain knowledge transfer methodology, which is expected to yield better synergy. 

In addition, the domain knowledge representation extracted from the teacher model must be defined. As we consider the language model of the transformer encoder series, we define the hidden states and attention matrices occurring in the encoder blocks of each layer as representations that imply domain knowledge. We distill the representation of English words observed in each input sequence $x_i$ from the teacher to the student model.

\begin{table*}[th!]
\centering
\resizebox{.9\hsize}{!}{%
\begin{tabular}{@{}lcccccccc@{}}
\toprule
 &
  \multicolumn{1}{l}{} &
  \textbf{Accuracy} &
  \textbf{AUROC} &
  \textbf{AUPRC} &
  \textbf{Recall} &
  \textbf{Precision} &
  \textbf{F1 Score} &
  \textbf{Average} \\ \midrule
\multicolumn{2}{c}{\textbf{Ko-BERT}}      & 71.9±0.2 & 78.5±0.1 & 82.8±0.3 & 76.2±0.2 & 75.8±0.2 & 76.0±0.2 & 76.9±0.2    \\
\multicolumn{2}{c}{\textbf{BERT-base}}     & 69.2±0.2 & 79.0±0.1 & 83.6±0.2 & 61.2±0.4 & \textbf{81.7±0.2} & 70.0±0.3 & 74.1±0.2          \\
\multicolumn{2}{c}{\textbf{KM-BERT}}       & 71.0±0.1 & 77.7±0.2 & 82.7±0.3 & 77.9±0.2 & 74.0±0.3 & 75.9±0.2 & 76.5±0.2          \\
\multicolumn{2}{c}{\textbf{Clinical-BERT}} & 70.9±0.1 & 77.6±0.1 & 82.7±0.2 & 77.1±0.2 & 74.3±0.3 & 75.6±0.1 & 76.4±0.2          \\
\multicolumn{2}{c}{\textbf{Bio-M-BERT}}    & 66.4±0.2 & 71.1±0.2 & 76.4±0.4 & 72.6±0.3 & 70.8±0.4 & 71.7±0.2 & 71.5±0.3          \\
\multicolumn{2}{c}{\textbf{RoBerta}}       & 70.6±0.2 & 77.3±0.2 & 81.7±0.2 & 71.0±0.2 & 77.1±0.3 & 73.9±0.1 & 75.3±0.2          \\
\multicolumn{2}{c}{\textbf{Bio-RoBerta}} &
  72.3±0.1 &
  78.4±0.1 &
  82.9±0.2 &
  79.3±0.2 &
  74.9±0.2 &
  77.0±0.1 &
  77.5±0.2 \\ \midrule
          & \multicolumn{1}{l}{}           & \multicolumn{7}{c}{\textbf{ours}}                                                                        \\
\multicolumn{1}{c}{\textbf{Student}} &
  \textbf{Teacher} &
  \multicolumn{1}{l}{} &
  \multicolumn{1}{l}{} &
  \multicolumn{1}{l}{} &
  \multicolumn{1}{l}{} &
  \multicolumn{1}{l}{} &
  \multicolumn{1}{l}{} &
  \multicolumn{1}{l}{} \\
\multicolumn{1}{c}{\textbf{Ko-Bert}} &
  \multicolumn{1}{l}{} &
  \multicolumn{1}{l}{} &
  \multicolumn{1}{l}{} &
  \multicolumn{1}{l}{} &
  \multicolumn{1}{l}{} &
  \multicolumn{1}{l}{} &
  \multicolumn{1}{l}{} &
  \multicolumn{1}{l}{} \\
          & \textbf{KM-BERT}               & 72.6±0.4 & 79.4±0.4 & 84.1±0.4 & \textbf{85.7±0.2} & 72.7±0.5 & \textbf{78.6±0.3} & \textbf{78.9±0.4} \\
          & \textbf{Clinical-BERT}         & 72.6±0.2 & 79.4±0.2 & 83.8±0.2 & 74.6±0.3 & 77.8±0.2 & 76.2±0.2 & 77.4±0.2          \\
          & \textbf{Bio-M-BERT}            & \textbf{73.4±0.2} & \textbf{80.4±0.2} & 84.3±0.2 & 75.1±0.3 & 78.6±0.2 & 76.8±0.2 & 78.1±0.2          \\
          & \textbf{RoBerta}               & 73.1±0.2 & 80.1±0.2 & {\ul 84.6±0.2} & 77.7±0.3 & 76.7±0.2 & 77.2±0.2 & 78.3±0.2          \\
          & \textbf{Bio-RoBerta}           & {\ul 73.3±0.2} & \textbf{80.4±0.2} & \textbf{85.0±0.2} & 77.6±0.3 & 76.9±0.2 & {\ul 77.3±0.2} & {\ul 78.4±0.2}    \\
\multicolumn{1}{c}{\textbf{Bert}} &
  \multicolumn{1}{l}{} &
  \multicolumn{1}{l}{} &
  \multicolumn{1}{l}{} &
  \multicolumn{1}{l}{} &
  \multicolumn{1}{l}{} &
  \multicolumn{1}{l}{} &
  \multicolumn{1}{l}{} &
  \multicolumn{1}{l}{} \\
          & \textbf{KM-BERT}               & 70.9±0.2 & 77.1±0.1 & 82.3±0.2 & 79.6±0.3 & 73.2±0.3 & 76.2±0.2 & 76.5±0.2    \\
          & \textbf{Clinical-BERT}         & 70.8±0.2 & 76.7±0.2 & 81.6±0.3 & 74.4±0.3 & 75.4±0.2 & 74.9±0.2 & 75.6±0.2          \\
          & \textbf{Bio-M-BERT}            & 70.4±0.1 & 76.5±0.2 & 81.6±0.2 & 77.3±0.2 & 73.6±0.2 & 75.4±0.2 & 75.8±0.2          \\
          & \textbf{RoBerta}               & 70.8±0.2 & 76.9±0.2 & 82.0±0.2 & {\ul 84.6±0.2} & 71.1±0.2 & 77.2±0.1 & 77.1±0.2 \\
          & \textbf{Bio-RoBerta}           & 68.6±0.2 & 76.3±0.2 & 81.2±0.3 & 64.6±0.4 & {\ul 80.0±0.3} & 70.7±0.3 & 73.2±0.3          \\ \bottomrule
\end{tabular}%
}
\caption[Performance table]{Performance table: This table presents the performances of different models in terms of all metrics. The mean and standard deviation values are listed. The models are validated with respect to a variety of BoW-based topic models (Random Forest (RF), Logistic Regression (LR), XGBoost, and Gradient Boosting (GB)) and Transformer-based models (Ko-BERT, BERT-base, KM-BERT, Clinical-BERT, Bio-M-BERT, RoBERTa, and Bio-RoBERTa). The column entitled "The Proposed Method" lists the results obtained using the proposed methodology---each combination of student and teacher models is evaluated. The metrics used for evaluation are accuracy, area AUROC, AUPRC, recall, precision, and F1 Score, and the average values are provided in the Average column. In the column corresponding to each metric, the best performance is highlighted in boldface and the second-best performance is underlined.}
\label{tab:Performance_tab}
\end{table*}

\subsubsection{Problem Formulation}
Let $S$ and $T$ contain $P$ transformer layers. Knowledge transfer is achieved by performing KD on the encoder layers of the student and teacher models. $S$ further solves the prediction task of classifying emergency/non-emergency cases, given $X$. Formally, the student receives domain knowledge from the teacher and solves the prediction task by minimizing the following objective function:

\begin{equation}
    \label{eq. Objective Function}
    \begin{aligned}
        \mathcal{L}_{total} = & \sum_{x \in X}\mathcal{L}_{pred}(S(x^S), Y) \\
        & + \sum_{d \in X }^{M} \sum_{p=0}^{P} \mathcal{L}_{layer}(f_p^S(d^S), f_p^T(d^T), \lambda)
    \end{aligned}
\end{equation}
where $\mathcal{L}_{pred}$ denotes the loss function that optimizes the classifier prediction of the student model performing the emergency/non-emergency classification task and $\mathcal{L}_{layer}$ denotes the loss function applied to a given layer of the model. Here, $f_p(\cdot)$ denotes the encoder block output in the $p$-th layer. The hyperparameter $\lambda$ comprises $\lambda = \{\alpha, \beta \}$, where $\alpha$ and $\beta$ are applied to $\mathcal{L}_{hidn}$ and $\mathcal{L}_{attn}$, respectively. 

\subsubsection{Knowledge distillation}
In this section, we mathematically express the process by which a student model receives domain knowledge from a teacher model. To this end, two types of KD are performed: hidden-state and attention-matrix distillation.\cite{sanh2019distilbert, jiao2019tinybert}

\textbf{Hidden-state Loss} is expressed as follows :
\begin{equation}
    \begin{aligned}
    \mathcal{L}_{hidn} = \alpha \mathsf{MSE}(\hat H^S, \hat H^T)
    \end{aligned}
\end{equation}
Let $H_S$ and $H_T$ be the hidden states of the student and teacher encoder layers, respectively. Let $H^S$ and $H^T$ be $H^S \in \mathbb{R}^{l \times d}$ and $H^T \in \mathbb{R}^{l' \times d'}$, respectively. (Note that $d = d',$ $l = l'$).

$\hat H^S$ and $\hat H^T$ are expressed as follows:
\begin{equation}
    \label{eq. domain knowledge hidden states}
    \begin{aligned}
    \hat H^S &= \sum_{i=0}^k\text{Avg.Pool}(H^S_{m^S_i}, m^S) \in \mathbb{R}^{k \times d} \\ 
    \hat H^T &= \sum_{i=0}^k\text{Avg.Pool}(H^T_{m^T_i}, m^T) \in \mathbb{R}^{k \times d'}
    \end{aligned}
\end{equation}
Let $m^S$ be a mask marking the domain knowledge words that the student model should receive from the teacher model. $m^T$ is a mask marking the domain knowledge words of the teacher model. $m^S \in \mathbb{R}^{l}$ and $m^T \in \mathbb{R}^{l'}$.

\textbf{Attention Matrices Loss} is expressed as follows :
\begin{equation}
    \begin{aligned}
    \mathcal{L}_{attn} = \beta \mathsf{MSE}(\hat A^S, \hat A^T)
    \end{aligned}
\end{equation}
Let $A_S$ and $A_T$ be the hidden states of the encoder layer for the student and teacher models, respectively. Let $A^S$ and $A^T$ be $A^S \in \mathbb{R}^{h \times l \times l}$ and $A^T \in \mathbb{R}^{h \times l \times l}$, respectively (but $l = l'$).

$\hat A^S$ and $\hat A^T$ are expressed as follows:
\begin{equation}
    \label{eq. domain knowledge attention matrices}
    \begin{aligned}
    \hat A^S &= \sum_{i=0}^k\text{Avg.Pool}(A^S_{m^S_i}, m^S) \in \mathbb{R}^{k \times l} \\ 
    \hat A^T &= \sum_{i=0}^k\text{Avg.Pool}(A^T_{m^T_i}, m^T) \in \mathbb{R}^{k \times l'}
    \end{aligned}
\end{equation}

Finally, for each encoder layer, the loss is calculated using the following formula:
\begin{equation}
    \begin{aligned}
    \mathcal{L}_{layer} = \begin{cases}     
        \alpha \mathcal{L}_{hidn} & \text{n = 0} \\
        \alpha\mathcal{L}_{hidn} + \beta\mathcal{L}_{attn} & {n \neq  0}
    \end{cases}
    \end{aligned}
\end{equation}
where $n=0$ denotes the embedding layer and $\mathcal{L}_{attn}$ is not performed because no attention matrix is generated.

\subsubsection{Prediction Loss}
The student model continues to perform and optimize the classification task.

\begin{equation}
    \begin{aligned}
    \mathcal{L}_{pred} = \mathsf{CE}(S(x^S), Y)
    \end{aligned}
\end{equation}

where $x^S$ denotes the input to the student model and $S(x^S)$ denotes the logits of the student model. $\mathsf{CE(\cdot)}$ denotes the cross entropy loss and $Y$ is the label.

\section{Experiments}
\label{sec:Experiments}

\subsection{Dataset}
The proposed methodology is validated using EMR data obtained from Korean PEDs. The task considered in this study is to classify the data into emergencies and non-emergencies using binary classification.

\subsection{Implementation Details}
In this section, we introduce the experimental environment and models on which the proposed methodology is applied. The A6000 GPU device is used, with a batch size of 32, 15 epochs of training with an early termination condition at 15 to choose the weight with the best valid performance. The learning rate is set to 1e-5 for the BERT body and 1e-2 for the classifier. For reproducibility, we set the random seed to 42.

The transformer-based language models Ko-BERT and BERT-base are used as baseline student models, and KM-BERT, Clinical-BERT (C-BERT), RoBerta, Bio-RoBerta (B-RoBerta), and Bio-Multilingual-BERT (Bio-M-BERT) are used as baseline teacher models.  In addition, four BoW-based machine learning models (RF, LR, XGBoost, and GB) are trained and compared.

The transformer-based models have $M$=12, $d$=768, $l$=512, and $h$=12, where $M$ denotes the number of encoder layers, $d$ denotes the dimension of hidden state, $l$ denotes the maximum sequence length, and $h$ denotes the number of multi-headers. The classification performances of the transformer models trained and measured on the test set are compared with that of the proposed method with all possible combinations of student-teacher models while training on the test set. In the proposed methodology, knowledge transfer and classification are performed simultaneously.

\subsection{Results}
Table ~\ref{tab:Performance_tab} presents a comparison of all methods in different groups in terms of accuracy, AUROC, AUPRC, recall, precision, and F1 Score. In addition, the average values are listed in the Average column to facilitate the overall performance comparison. In general, transformer-based NLP models are observed to outperform traditional NLP processes proposed in previous studies. Transformer-based language models can adapt quickly to the language patterns present in data when fine-tuned using language patterns during pretraining. 

In particular, it is interesting to note from Table~\ref{tab:Performance_tab} that language models pre-trained with data from the medical field (KM-BERT, Clinical-BERT, etc.) perform worse than general language models (BERT-based, Ko-BERT) in terms of AUROC and AUPRC. This is because, as explained earlier, general language models perform favorably with respect to N-lingual and unstructured data while pretraining with a large amount of data, whereas language models pre-trained on the medical domain struggle on N-lingual or unstructured real data, even though they have learned medical domain information.

The bottom part of Table~\ref{tab:Performance_tab} presents the results obtained by training the existing language models using the proposed methodology. The student models are divided into Ko-BERT- and BERT-based cases and teacher models are taken to be pre-trained language models in the medical domain. The results are observed to be significantly better than those obtained by fine-tuning a simple language model. Moreover, the performance is even better than that of existing student and teacher models when trained alone. The aforementioned experimental results demonstrate the effectiveness of knowledge transfer between language models, leading to efficient interaction between knowledge of different language models.

\subsubsection{Qualitative analysis}
\begin{figure}[h!]
\centering
\includegraphics[width=\columnwidth]{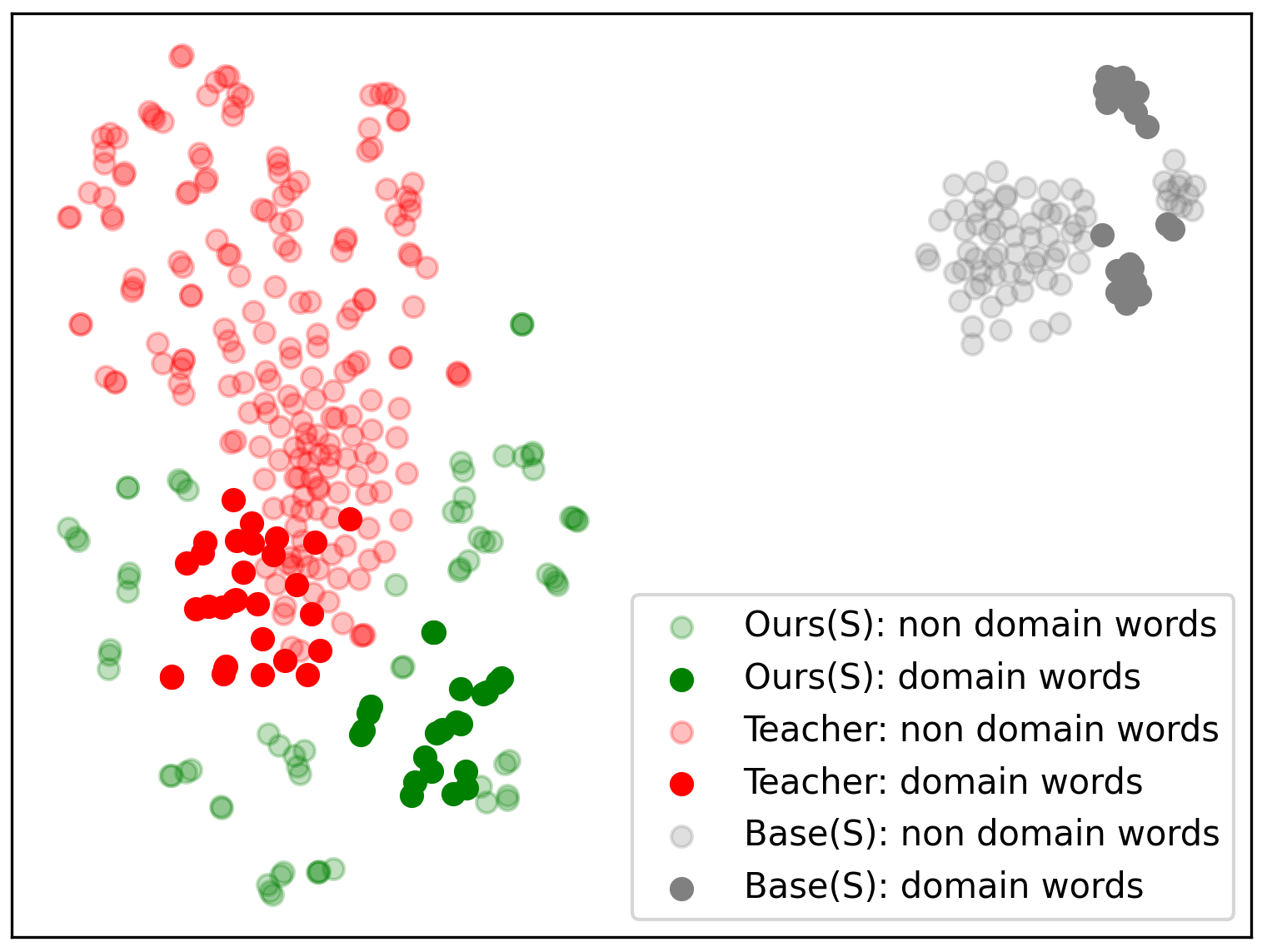} 
\caption{Qualitative analysis: Visual representation of the effectiveness of the proposed training method. S denotes a student model, which is taken to be Ko-BERT, and the teacher is taken to be KM-BERT. The gray dots correspond to an independently trained student model, the green dots correspond to the proposed training methodology, and the red dots correspond to the spatial coordinates of the teacher model's representation}
\label{fig:Qualitative_analysis}
\end{figure}

To verify the relationship between domain words and teacher embeddings before and after DSG-KD visually, we present a scatter plot in Figure ~\ref{fig:Qualitative_analysis}. Random samples are obtained from the dataset and visualized. In the figure, Ko-BERT is used as the student model and KM-BERT is used as the teacher model.

First, for each color, the darker shade represents the coordinates of the embedding vectors for words that contain domain knowledge, and the lighter shade represents the coordinates of the embedding vectors for words that do not. The gray points represent the coordinates of the embedding space when the student model is trained independently. The green points represent the coordinates of the embedding space of the student model trained using the proposed methodology. The red points represent the coordinates of the embedding space corresponding to the teacher model, which provides a good understanding of the medical domain. Visually, the embedding space of the student model trained using the proposed methodology is similar to that of the teacher model, which is rich in medical expressions. In contrast, when the student model is trained independently, it yields a relatively poor representation space and its visual representation is located far from that of the teacher model.

\subsubsection{Case analysis}
\begin{figure}[h!]
\centering
\includegraphics[width=\columnwidth]{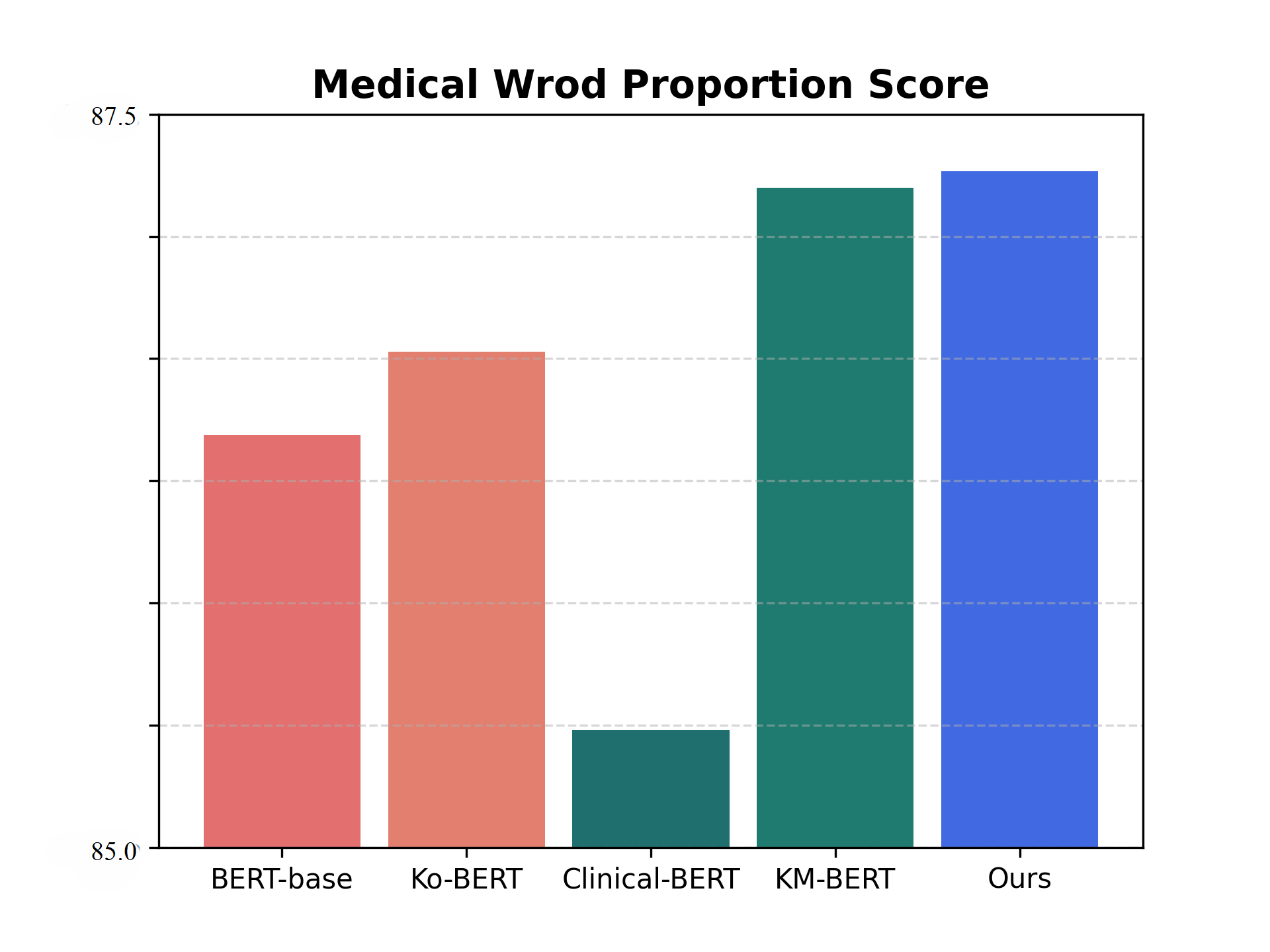} 
\caption{Case analysis: This figure shows calculated MWPS for correctly classified cases in the test set for each model. The proposed model (Ours) uses Ko-BERT as the student model and KM-BERT as the teacher model. By achieving the highest MWPS performance compared to all other models, our methodology demonstrates an effective understanding of medical words. Notably, the higher MWPS compared to the teacher model is particularly encouraging in terms of domain knowledge comprehension.}
\label{fig:fail_check}
\end{figure}

Furthermore, to understand the impact of the proposed methodology on Domain words (in this study, medical words), we defined and measured the Medical Word Proportion Score (MWPS). The MWPS is defined as follows:

\begin{equation}
    \begin{aligned}
    \mbox{MWPS} = \frac{\mbox{\% of medical terms in English words}}{\mbox{\% of English words in all words}}
    \end{aligned} 
\end{equation}
This can be expressed mathematically as:
\begin{equation}
    \begin{aligned}
    \text{MWPS} = \frac{\frac{m}{E}}{\frac{E}{A}} = \frac{m \cdot A}{E^2}
    \end{aligned}
\end{equation}
where $m$ is the number of medical terms in English words, $E$ is the total number of English words, and $A$ is the total number of words.

We measured the MWPS for correctly classified cases in the test set for each model. This allowed us to evaluate the effectiveness of the proposed methodology. The comparison of MWPS for each model is shown in Figure~\ref{fig:fail_check}. From the figure, we can observe that our model achieves the highest MWPS, indicating that the DSG-KD methodology effectively learns Domain words, i.e., Domain Knowledge. Specifically, it demonstrates that medical knowledge can be transferred to a general language model and that our model shows a slightly higher understanding of medical terms than KM-BERT. In summary, this suggests that the proposed methodology is highly beneficial for applications in specific domains such as the medical field. Thus, it is proven that the DSG-KD methodology learns the medical knowledge embedded in medical terms and achieves superior classification performance.

\subsection{Ablation Study}
The influence of $\mathcal{L}_{hidn}$ and $\mathcal{L}_{attn}$ on performance is investigated by considering Equation 8. The influences of $\alpha$ and $\beta$ are also studied. Two additional experiments are performed with Ko-BERT as the student model and KM-BERT as the teacher model. The performances are evaluated in terms of three metrics: AUROC, AUPRC, and F1 Score.

\subsubsection{Influence of $\mathcal{L}_{hidn}$ \& $\mathcal{L}_{attn}$ on the objective function}

Table ~\ref{tab:Ablation_1} presents the effects of $\mathcal{L}_{hidn}$ and $\mathcal{L}_{attn}$ on $\mathcal{L}_{total}$. The baseline model, which includes all loss components, exhibits an AUROC of 79.4±0.4, an AUPRC of 84.1±0.4, and an F1 Score of 78.6±0.3. This is used as the baseline for comparison in the ablation study, which is performed by removing $\mathcal{L}_{hidn}$ and $\mathcal{L}_{attn}$ individually, or in combination. The results are as follows. 

Case 1: Removing both $\mathcal{L}_{hidn}$ and $\mathcal{L}_{attn}$: When $\mathcal{L}_{hidn}$ and $\mathcal{L}_{attn}$ are removed, all three performance metrics decrease, indicating that the two losses play significant roles in improving model performance. 
Case 2: Removing only $\mathcal{L}_{hidn}$: The absence of $\mathcal{L}_{hidn}$ also results in a decrease in performance, although to a lesser extent than in Case 1. This suggests that hidden losses contribute to model performance to a certain extent. 
Case 3: Removing $\mathcal{L}_{attn}$: The absence of $\mathcal{L}_{attn}$ has the largest effect on F1 Score, suggesting that this factor is particularly important for the accuracy and recall balance of the model. In contrast to Cases 2 and 3, removing both $\mathcal{L}_{hidn}$ and $\mathcal{L}_{attn}$ further degrades the performance, emphasizing the importance of synergistic functioning of these components to achieve high model performance.

This study demonstrates that each component plays an important role, and that prediction loss is the most important factor in maintaining the integrity of model performance. The cumulative degradation caused by removing multiple components suggests that the interactions among these components are complex and essential for high-quality model learning.

\begin{table}[t]
\centering
\resizebox{\columnwidth}{!}{%
\begin{tabular}{@{}cccccc@{}}
\toprule
 \multirow{2}{*}{$\bm{\mathcal{L}_{pred}}$} &
  \multirow{2}{*}{$\bm{\mathcal{L}_{hidn}}$} &
  \multirow{2}{*}{$\bm{\mathcal{L}_{attn}}$} &
  \multirow{2}{*}{\textbf{AUROC}} &
  \multirow{2}{*}{\textbf{AUPRC}} &
  \multirow{2}{*}{\textbf{F1 Score}} \\
                    &                     &                     &                      &                      &                      \\ \midrule
\textbf{\checkmark} & \textbf{\checkmark} & \textbf{\checkmark} & \textbf{79.4±0.4} & \textbf{84.1±0.4} & \textbf{78.6±0.3} \\
\textbf{\checkmark} & \textbf{-}          & \textbf{-}          & 78.5±0.1          & 82.8±0.3          & 76.0±0.2          \\
\textbf{\checkmark} & \textbf{\checkmark} & \textbf{-}          &  {\ul{79.2±0.3}}    &  {\ul{83.9±0.3}}    & {\ul{78.6±0.2}}    \\
\textbf{\checkmark} & \textbf{-}          & \textbf{\checkmark} & 78.1±0.3          & 83.1±0.4          & 67.8±0.3          \\ \bottomrule
\end{tabular}%
}
\caption{Ablation study on objective function for $\mathcal{L}_{hidn}$, $\mathcal{L}_{attn}$}
\label{tab:Ablation_1}
\end{table}


\subsubsection{Effect of $\alpha$ \& $\beta$ on the objective function}
Table ~\ref{tab:Ablation_2} presents the impact of $\alpha$ and $\beta$ on the performance, expressed in terms of performance. When $\alpha = 0.6$ and $\beta = 0.2$, AUROC = 79.4±0.4, AUPRC = 84.1±0.4, and F1 Score = 78.6±0.3 are observed. We systematically vary $\alpha$ and $\beta$ to observe changes in the performance, and the results are presented below.

Case 1: Impact of equal weights ($ \alpha = \beta = 1.0 $): Assigning equal weights to both hyperparameters results in a slight decrease in all metrics, suggesting that the model may not need to emphasize the factors controlled by $\alpha$ and $\beta$ equally. 
Case 2: Increasing the emphasis on $\beta$ ($\alpha = 1.0$, $\beta = 0.6$): Increasing $\beta$ relative to $\alpha$ slightly improves the AUROC to 79.0±0.4, indicating that aspects of the model influenced by $\beta$ may be more important for predictive ability. 
Case 3: Gradual adjustment of $\alpha$ and $\beta$: A pattern is observed when proportionally adjusting $\beta$ from 1.0 to 0.1 while gradually decreasing $\alpha$ from 1.0 to 0.1, with the highest F1 Score achieved when $\alpha = 0.8$ and $\beta = 0.4$, and the best AUROC and AUPRC at $\alpha = 0.8$ and $\beta = 0.4$. This suggests an optimal range for the hyperparameters that balances the factors controlled by the model.
Case 4: Significant reduction of $\alpha$ ($\alpha = 0.1$, $\beta = 0.1$): Significantly reducing both $\alpha$ and $\beta$ to 0.1 results in the lowest AUROC, a slight increase in AUPRC, and a decrease in the F1 Score, which suggests that very low values of these hyperparameters can degrade the performance of the model, especially in terms of precision and recall balance.

In conclusion, the hyperparameters $\alpha$ and $\beta$ play important roles in the performance of the Ko-BERT + KM-BERT model. For each data environment, an optimal parameter range yields the best performance. This emphasizes the importance of fine-tuning hyperparameters to satisfy specific performance goals during model training.

\begin{table}[t]
\centering
\resizebox{.8\hsize}{!}{%
\begin{tabular}{@{}ccccc@{}}
\toprule
{$\bm{\alpha}$}   & {$\bm{\beta}$}   & \textbf{AUROC} & \textbf{AUPRC} & \textbf{F1 Score} \\ \midrule
\textbf{1.0} & \textbf{1.0} & 78.8±0.4    & 83.4±0.4    & 78.2±0.3       \\
\textbf{1.0} & \textbf{0.6} & 79.0±0.4    & 83.5±0.4    & 75.1±0.4       \\
\textbf{0.9} & \textbf{0.5} & 78.5±0.4    & 83.1±0.4    & 78.3±0.3       \\
\textbf{0.8} & \textbf{0.4} & 79.0±0.4    & 83.6±0.4    & {\ul{78.4±0.3}} \\
\textbf{0.7} & \textbf{0.3} & {\ul{79.1±0.4}}    & {\ul{83.6±0.3}}    & 77.9±0.4          \\
\textbf{0.6} & \textbf{0.2} & \textbf{79.4±0.4} & \textbf{84.1±0.4} & \textbf{78.6±0.3} \\
\textbf{0.1} & \textbf{0.1} & 77.8±0.3    & 81.7±0.5    & 76.9±0.3       \\ \bottomrule
\end{tabular}%
}
\caption[Ablation 2]{Ablation study on objective function for $\alpha$ and  $\beta$}
\label{tab:Ablation_2}
\end{table}

\section{Conclusions}
This study addresses the need for effective domain knowledge transfer in pre-trained language models and provides a promising approach for bridging the gap between general-purpose language understanding and domain-specific expertise. As the demand for domain expertise in NLP continues to increase, this study makes a timely and valuable contribution to improving the ability of language models to utilize specific and nuanced domain knowledge.

To this end, we introduce a novel methodology to extract domain knowledge from domain-specific pre-trained language models and transfer the knowledge to more generalized language models. We find that domain-specific pre-trained language models are vulnerable to bilingual data in non-English-speaking countries such as Korea, and demonstrate that by transferring domain knowledge to generalized language models, the proposed methodology is robust in N-lingual environments and exhibits effective classification performance in specific domains.

In this study, a highest average metric of 789±0.4 is achieved by transferring medical knowledge from the teacher model (KM-BERT) to the student model (Ko-BERT) using Korean emergency room EMR data. This demonstrates that the proposed framework can be used as a tool to assist doctors and healthcare workers in decision-making by helping them overcome N-lingual challenges inherent in EMR data recorded in non-English speaking countries. Beyond the medical field, the methodology can also be applied in various professional and technical fields, opening up potential avenues for its use as a decision aid in many forms.

In future works, we intend to extend and improve this methodology by utilizing EMR data obtained from diverse non-English-speaking countries, or by considering downstream tasks from various technical and professional fields in specific non-English-speaking countries. We also aim to develop an advanced model architecture by applying the latest KD techniques to deep learning methods.

\bibliographystyle{unsrt}
\bibliography{reference}

\EOD

\end{document}